\documentclass{article}
\usepackage{ijcai23}

\usepackage{times}
\usepackage{soul}
\usepackage{url}
\usepackage[hidelinks]{hyperref}
\usepackage[utf8]{inputenc}
\usepackage[small]{caption}
\usepackage{graphicx}
\usepackage{amsmath}
\usepackage{amsthm}
\usepackage{booktabs}
\usepackage{bm}
\usepackage{pifont}
\usepackage{amssymb}
\usepackage{color}
\usepackage[switch]{lineno}

\title{Semantic-aware Generation of Multi-view Portrait Drawings}

\author{
	Biao Ma$^1$
	\and
	Fei Gao$^{2,3*}$\and
	Chang Jiang$^1$\and
	Nannan Wang$^4$\and
	Gang Xu$^{1}$
	\affiliations
	$^1$ School of Computer Science and Technology, Hangzhou Dianzi University\\
	$^2$ Hangzhou Institute of Technology, Xidian University; $^3$ AiSketcher Tech.\\
	$^4$ ISN State Key Laboratory, Xidian University\\
	\emails
	\{aiartma, jc233, gxu\}@hdu.edu.cn,
	\{fgao, nnwang\}@xidian.edu.cn
}

\begin{document}

\maketitle


\begin{abstract}
Neural radiance fields (NeRF) based methods have shown amazing performance in synthesizing 3D-consistent photographic images, but fail to generate multi-view portrait drawings. 
The key is that the basic assumption of these methods -- \textit{a surface point is consistent when rendered from different views} -- doesn't hold for drawings. 
In a portrait drawing, the appearance of a facial point may changes when viewed from different angles. 
Besides, portrait drawings usually present little 3D information and suffer from insufficient training data. 
To combat this challenge, in this paper, we propose a \textit{Semantic-Aware GEnerator} (SAGE) for synthesizing multi-view portrait drawings. 
Our motivation is that facial semantic labels are view-consistent and correlate with drawing techniques.
We therefore propose to collaboratively synthesize multi-view semantic maps and the corresponding portrait drawings. 
To facilitate training, we design a semantic-aware domain translator, which generates portrait drawings based on features of photographic faces. 
In addition, use data augmentation via synthesis to mitigate collapsed results.
We apply SAGE to synthesize multi-view portrait drawings in diverse artistic styles. Experimental results show that SAGE achieves significantly superior or highly competitive performance, compared to existing 3D-aware image synthesis methods. The codes are available at \url{https://github.com/AiArt-HDU/SAGE}. 
\end{abstract}


\section{Introduction}
\label{sec:intro}
%

\begin{figure}
	\begin{center}
		\includegraphics[width=1\linewidth]{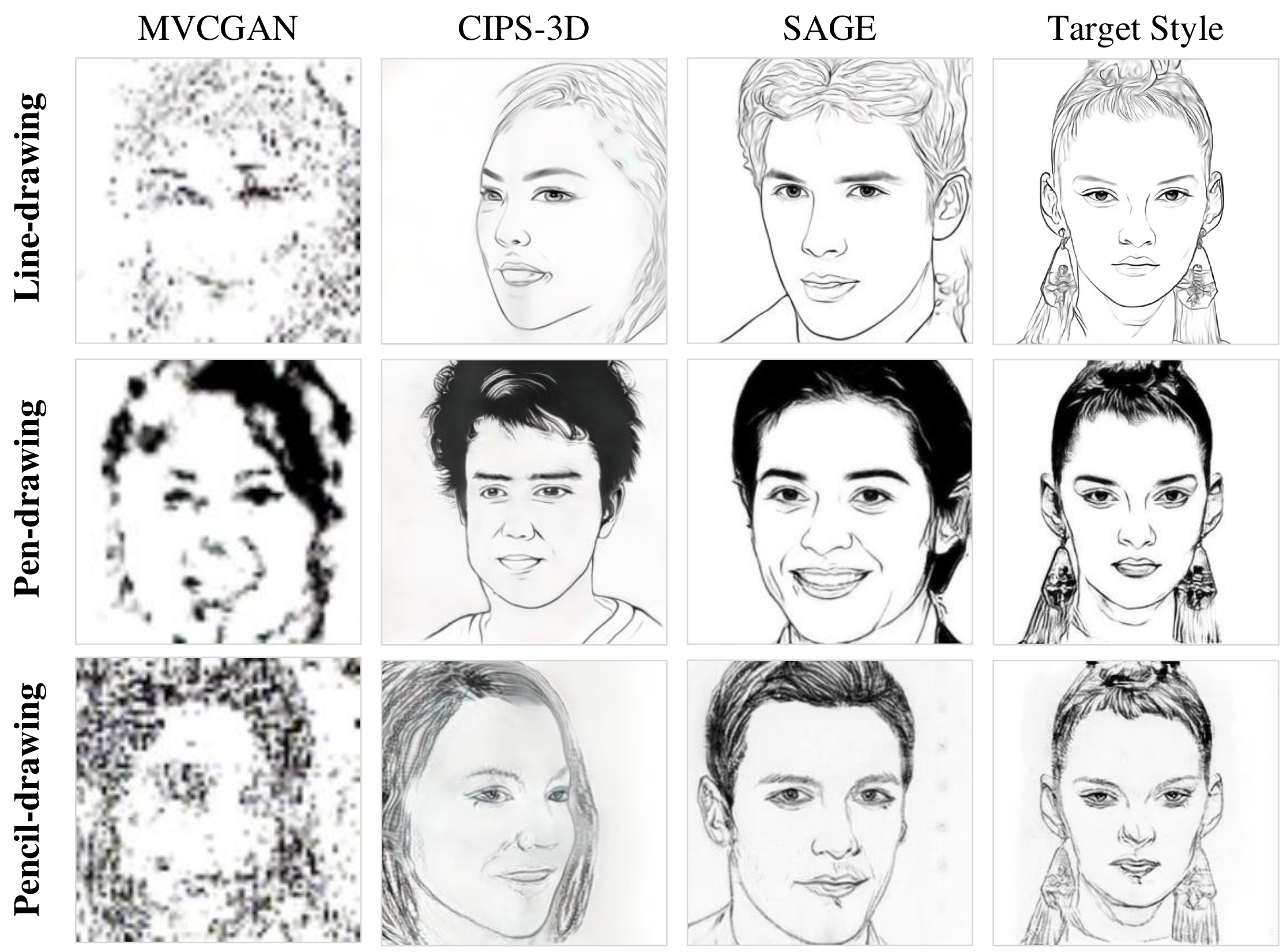}
	\end{center}
	 \vspace{-0.4cm}
	\caption{Portrait drawings synthesized by MVCGAN \protect\cite{zhang2022mvcgan}, CIPS-3D \protect\cite{zhou2021cips}, and our method, i.e. SAGE. The final column show training examples in target styles. 
	}
	 \vspace{-0.4cm}
	\label{fig:mtv}
\end{figure}

3D-aware image synthesis \cite{mildenhall2021nerf} aims to generate multi-view consistent images and, to a lesser extent, extract 3D shapes, without supervision on geometric or multi-view image datasets. 
Recently, inspired by the great success of Neural Radiation Fields (NeRF) \cite{mildenhall2021nerf} and Generative Adversarial Networks (GANs) \cite{karras2019style}, impressive progress has been achieved in generating multi-view photos as well as detailed geometries \cite{deng2022gram,xiang2022gram,chan2022efficient}. 
Besides, several recent methods \cite{zhang2022mvcgan,gu2021stylenerf,niemeyer2021giraffe,zhou2021cips} can also synthesize high quality artistic images, such as oil-paintings. 
While marveling at the impressive results of 3D-aware image synthesis methods, we wish to extend the style of synthesized images. 
Unfortunately, the advanced methods all fail to generate high quality multi-view portrait drawings, e.g. facial line-drawings (Fig. \ref{fig:mtv}).

\begin{figure}[t]
	\begin{center}
		\includegraphics[width=0.75\linewidth]{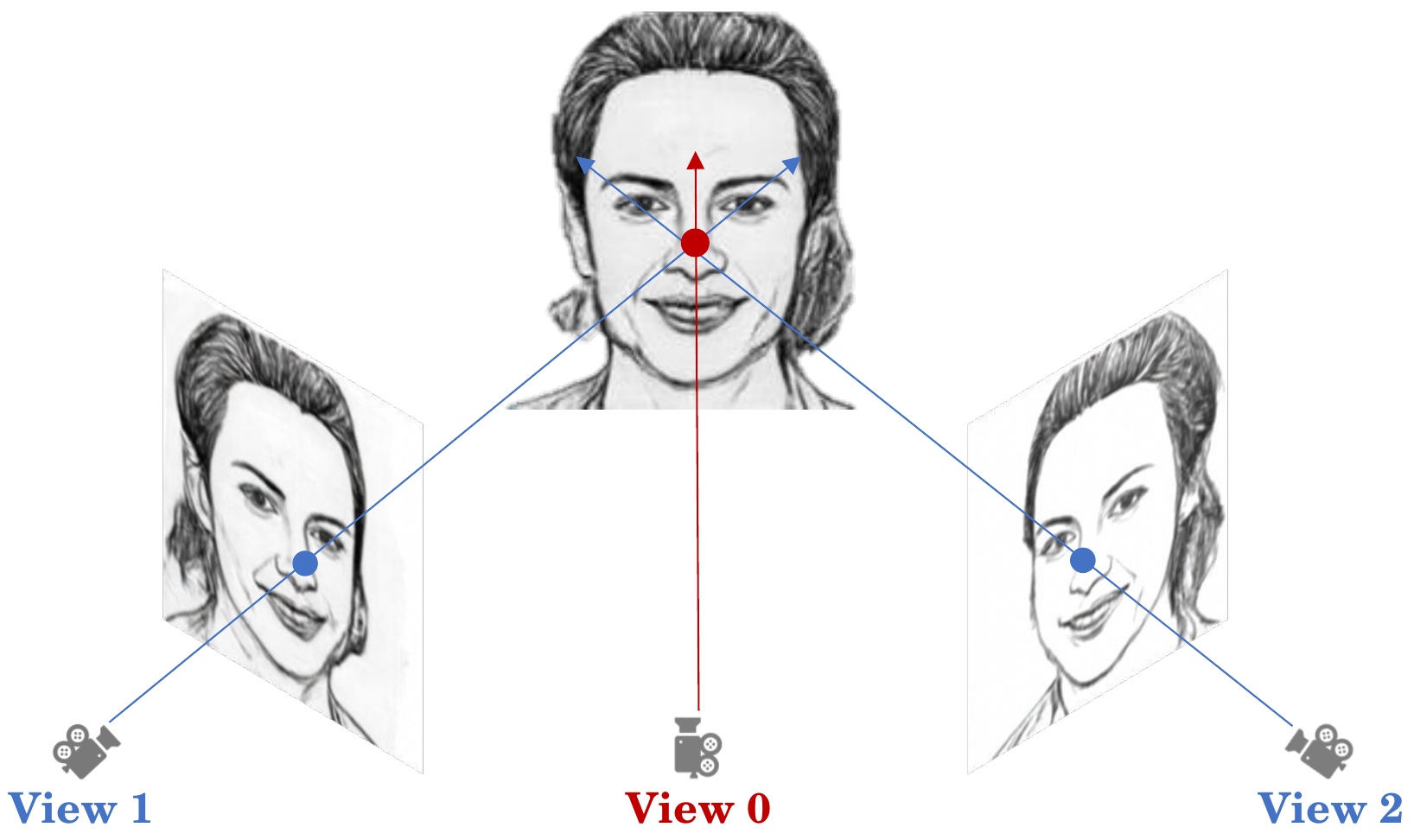}
	\end{center}
	 \vspace{-0.3cm}
	\caption{
	In portrait drawings, the appearance of a facial point may be inconsistent between different views. For example, the nose tip is represented by strokes in both View 1 and View 2, but is left blank in View 0. 
	}
	 \vspace{-0.5cm}
	\label{fig:sketchview}
\end{figure}

There are mainly three reasons for their failure.
First, the assumption of NeRF-based methods -- \textit{a surface point is consistent when rendered from different views} -- doesn't hold for drawings. 
Human artists typically use a sparse set of strokes to represent geometric boundaries, and use diverse levels of tones to present 3D structures \cite{sousa1999pencil}. 
As illustrated in Fig. \ref{fig:sketchview}, the boundaries may vary when a face is viewed from different angles.
In other words, the appearance of a facial point may be inconsistent between different views, in portrait drawings. 
Second, portrait drawings usually present sparse information with little 3D structures. Existing NeRF-based methods produce radiance fields and render images based on adjacent correlations and stereo correspondence \cite{zhang2022mvcgan,deng2022gram}. As a result, it is not appropriate to directly apply previous methods for portrait drawing synthesis. 
Third, previous methods require a large amount of training data. Unfortunately, it is extremely time-consuming and laborious for human artists to create adequate portrait drawings. 



To combat this challenge, in this paper, we propose a \textit{Semantic-Aware GEnerator} (SAGE) for synthesizing multi-view portrait drawings. 
Our motivation is that facial semantic labels are view-consistent and are highly correlated with the appearance of portrait drawings. Commonly human artists draw different semantic areas by using adaptive drawing techniques \cite{sousa1999pencil}. 
We therefore collaboratively synthesize multi-view consistent semantic maps and the corresponding portrait drawings. 
Besides, we use semantic maps to guide synthesis of portrait drawings, through semantic-adaptive normalization \cite{wang2018high}. 
As a result, the synthesized drawings are constrained to convey facial semantic structures, instead of multi-view consistency. 

In addition, NeRF-based modules, including radiation fields production and volume rendering (VR) \cite{zhang2022mvcgan}, are still essential for producing multi-view images.
For effective learning of such modules, we propose a two-stage generation and training strategy to make NeRF-based modules suitable for synthesizing portrait drawings.
First, we train the NeRF modules for generating multi-view facial photos and semantic maps in parallel.
Afterwards, we use a semantic-adaptive domain translator to synthesizes portrait drawings from features of photographic faces. 
The warm-up in the first stage makes NeRF-based methods suitable for synthesizing portrait drawings. Consequently, the final generator becomes capable of producing high quality portrait drawings under different views.
Finally, we use data augmentation via synthesis to obtain adequate training samples, and to mitigate collapsed results under large pose variants.

We apply our method, SAGE, to synthesize diverse styles of portrait drawings, including pen-drawings \cite{yi2019apdrawinggan}, line-drawings, pencil-drawings \cite{Fan2022FS2K}, and oil-paintings \cite{nicholpainter}. Experimental results show that SAGE achieves significantly superior or highly competitive performance, compared to existing 3D-aware image synthesis methods. Especially, our method stably generate high quality results across a wide range of viewpoints. 

\section{Related Work}
\label{sec:related}
\subsection{3D-aware Image Synthesis}
\label{ssec:3daware}
3D-aware image synthesis aims to explicitly control the camera view of synthesized images. Recently, numerous NeRF-based methods have been proposed and achieved impressive progress. For example, to achieve more photorealistic results, StyleNeRF \cite{gu2021stylenerf} uses StyleGAN \cite{karras2019style} to obtain high-resolution images by up-sampling low-resolution feature maps through Convolutional Neural Networks (CNNs). StyleSDF \cite{or2022stylesdf} employs a signed distance field to obtain low-resolution spatial features and then produces high-resolution images through up-sampling. GIRAFFE \cite{niemeyer2021giraffe} instead up-samples multi-scale feature maps based on GRAF \cite{schwarz2020graf} to generate high-resolution and multi-view consistent images. Recently, MVCGAN \cite{zhang2022mvcgan} introduces a method to warp images based on camera matrices, based on pi-GAN \cite{chan2021pi}. GRAM \cite{deng2022gram,xiang2022gram} optimizes point sampling and radiance field learning. 

A similar work to ours is CIPS-3D \cite{zhou2021cips}, which use a NeRF-based network for view control and a 2D implicit neural representation (INR) network for generating high-resolution images. CIPS-3D is also trained in two-stages so as to synthesize artistic portraits, e.g. oil-paintings and cartoon faces. However, CIPS-3D produces geometric deformations and unpleasant artifacts, when applied to portrait drawings (Fig. \ref{fig:mtv}). 
Differently, we propose a novel two-stage generation framework, i.e. using a domain translator to synthesize portrait drawings conditioned on photographic features. Besides, we generate multi-view images based on semantic consistency instead of 3D consistency. 
Recently, both FENeRF \cite{sun2022fenerf} and IDE-3D \cite{sun2022ide} directly use NeRF to render facial semantics, as well as multi-view images, for face editing. Differently, we decode both facial semantics and portrait drawings from photographic features, and use semantics to provide structural guidance. 


\begin{figure*}
	\begin{center}
		\includegraphics[width=1.0\linewidth]{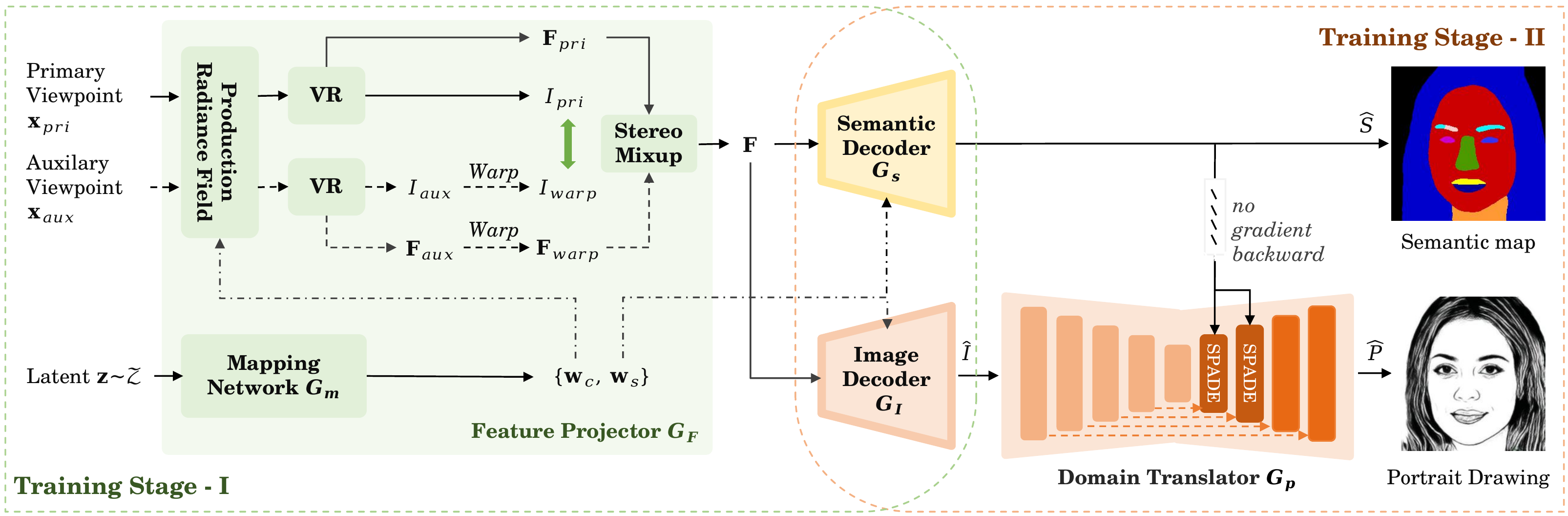}
	\end{center}
	\vspace{-0.3cm}
	\caption{Pipeline of our semantic-aware portrait drawing generator, SAGE. The feature projector $G_F$ generates feature maps $\mathbf{F}$ and normalization parameters $\mathbf{w}$, which control content and viewpoints of generated faces. 
In Stage-I, we train the feature projector $G_F$, semantic decoder $G_s$, and image decoder $G_I$ to enable the generator producing multi-view facial photos and semantic masks. 
In Stage-II, we add the portrait drawing decoder $G_p$, and refine all decoders for synthesizing high quality portrait drawings based on features of photographic faces. }
	\label{fig:pipline}
	\vspace{-0.2cm}
\end{figure*}

\subsection{2D Artistic Portrait Drawing Generation}
\label{ssec:apdg}
For now, there have been a huge number of methods for generating 2D artistic portrait drawings. 
Most existing methods try to translate facial photos to sketches, e.g. pencil-drawings \cite{Wang2009Face,Wang2017Face}, pen-drawings \cite{yi2019apdrawinggan}, and line-drawings \cite{gao2020making,YiLLR22}. The advanced methods are mainly based on 2D conditional GANs \cite{isola2017image,Gui2020ReviewGAN} and formulate this task as image-to-image translation. Researchers have made great efforts to boost the identity-consistency and texture-realism, by designing new learning architectures \cite{Zhang2019TIP,zhu2021sketch,Zhang2018IJCAI}, or using auxiliary geometric information, e.g. facial landmarks \cite{yi2019apdrawinggan,gao2020making}, semantics \cite{gao2021cagan,Li2021GENRE}, and depth \cite{chan2022informativedraw}. 
There are also several unsupervised methods for synthesizing 2D portrait drawings. These methods \cite{noguchi2019image,li2020few,ojha2021few,kong2021smoothing} aims at solving the few-shot learning problem of big generative models (e.g., StyleGAN series \cite{karras2019style}) and verify them on portrait drawings.
In this paper, we use a modification of GENRE \cite{Li2021GENRE} to synthesis 2D portrait drawings for data augmentation (Section \ref{ssec:dataaug}). Besides, the architecture of our domain translator is similar to GENRE (Section \ref{ssec:sage}).

\section{Method}
\label{sec:method}

Our whole model follows the architectures of NeRF-based GANs \cite{zhang2022mvcgan}. As shown in Fig. \ref{fig:pipline}, our generator $G$ includes four parts: a NeRF-based feature projector $G_F$, a semantic decoder $G_s$, and an image decoder $G_I$ followed by a domain translator $G_p$. $G_p$ produces portrait drawings conditioned on input latent code $\mathbf{z}$ and viewpoint $\mathbf{x}$. Correspondingly, we use three discriminators to judge realism of semantic maps, facial images, and drawings, respectively. 

\subsection{NeRF-based Feature Projection}
\label{ssec:nerf}
We first project latent codes and viewpoints to representations of multi-view faces by NeRF. 
Specially, we map the latent code $\mathbf{z}$ to parameters $\mathbf{w} = \{ \mathbf{w}_c, \mathbf{w}_s \}$ through a mapping network, $G_m$. $\textbf{w}_c$ controls content (e.g. identity, structure, and appearance) of the generated images. $\textbf{w}_s$ controls the style of synthesized portrait drawing. 
Given a viewpoint $\mathbf{x}$ in space, we map it to appearance parameters through a multilayer FiLM network \cite{perez2018film}. Afterwards, we use the \emph{Volume Rendering} (VR) \cite{deng2022depth} module  to produce a facial image and the corresponding feature map. 
In the implementation, $\textbf{w}_c$ is composed of multiple vectors representing frequencies and phases, and is fed into the FiLM layers. 
$\textbf{w}_s$ modulate deep features in sematic and image decoders, in the manner of AdaIN \cite{huang2017arbitrary}.

During training, we follow the settings of MVCGAN \cite{zhang2022mvcgan}. Specially, we render two images from primary viewpoint $\mathbf{x}_{pri}$ and auxiliary viewpoint $\mathbf{x}_{aux}$, respectively. The corresponding images, $I_{pri}$ and $I_{aux}$, represent the same face but with different poses. Let $\mathbf{F}_{pri}$ and $\mathbf{F}_{aux}$ be the corresponding feature representations. Afterwards, $I_{aux} / \mathbf{F}_{aux}$ are geometrically aligned to $I_{pri} / \mathbf{F}_{pri}$ thorough warping. 
The warped image $I_{warp}$ is constrained to be the same as $I_{pri}$ by a image-level reprojection loss (Eq.\ref{eq:Gstage1}).
Besides, the primary feature $\mathbf{F}_{pri}$ is linearly mixed with the warped feature $\mathbf{F}_{warp}$. The resulting mixed feature $\mathbf{F}$ is fed into the following decoders. 
In the testing stage, the auxiliary position and the stereo mixup module aren't needed. 


\subsection{Semantic-aware Portrait Generation}
\label{ssec:sage}
To enable the network successfully produce facial structures, we propose to decode portrait drawings from features of photographic faces. 
Besides, we use 2D semantic information to guide the synthesis of portrait drawings. 

\begin{figure}
	\begin{center}
		\includegraphics[width=0.9\linewidth]{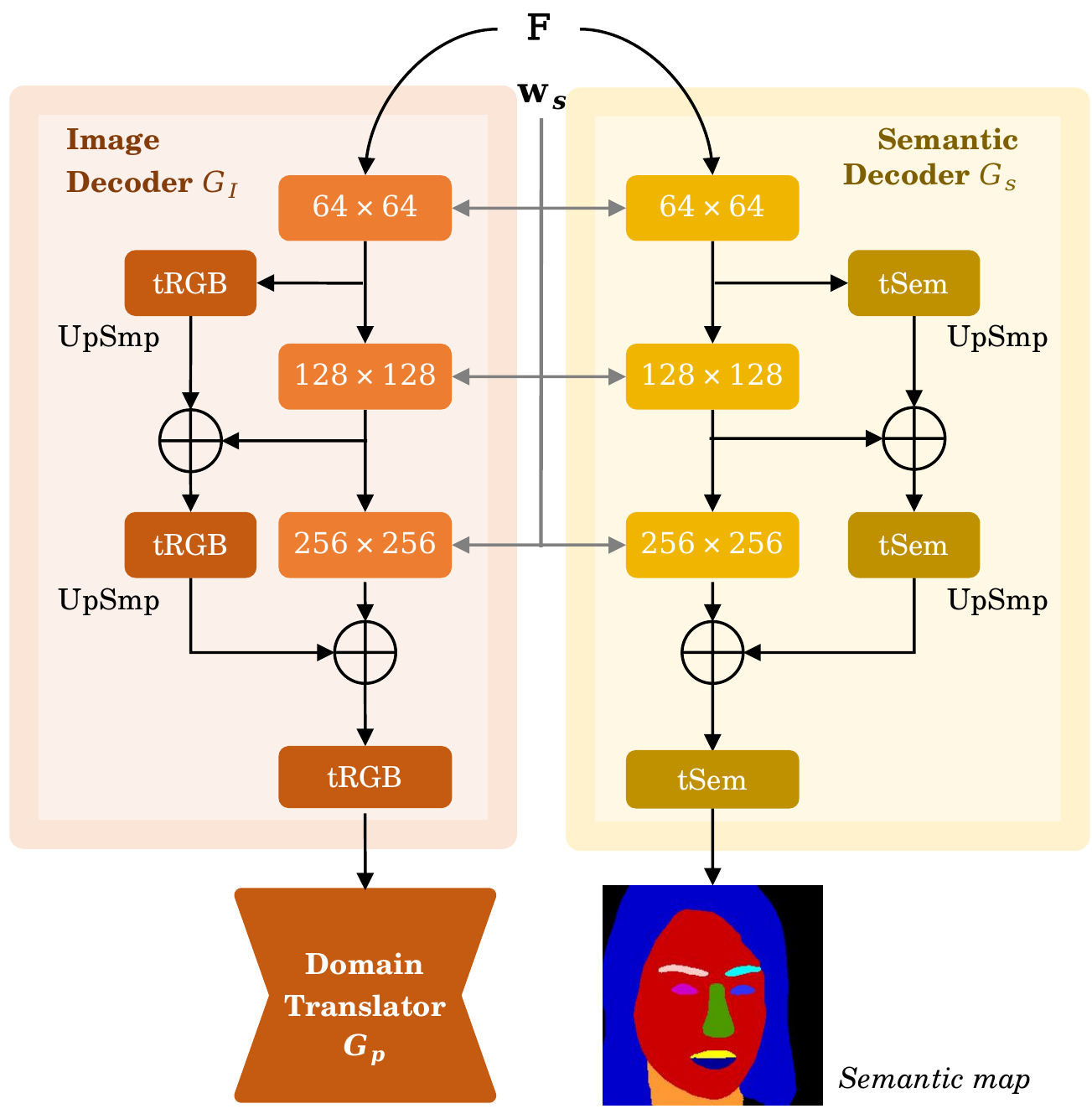}
	\end{center}
	 \vspace{-0.35cm}
	\caption{Semantic decoder (right) and portrait decoder (left and bottom). The $\mathrm{tRGB}$ and $\mathrm{tSem}$ module map deep features to a RGB image and semantic maps by $1\times 1$ convolution, respectively. $\mathrm{UpSmp}$ denotes upsampled interpolation.}
	\label{fig:decoder}
	\vspace{-0.5cm}
\end{figure}

\textbf{Semantic and Image Decoders.}
First, we use a semantic $G_s$ and an image decoder $G_I$ to collaboratively synthesize semantic maps and facial photos.  
Both decoders follow the same architectures, but with different numbers of output channels. 
As shown in Fig. \ref{fig:decoder}, each decoder contains three upsampling convolutional layers, and progressively generates high-resolution outputs.
As previously mentioned, features over each layer are channel-wisely modulated by $\mathbf{w}_s$ in the manner of AdaIN. 
Besides, we produce multi-scale outputs and integrate them together for producing a final output.  
The image decoder, $G_I$, generates a RGB facial image, i.e. 
\begin{equation}
		\hat{I}^{3\times 256\times 256} = G_I(\mathbf{F}, \textbf{w}_{s}).
	\label{eq:I}
\end{equation}
The semantic decoder, $G_s$, produces the corresponding 19-channel semantic masks \cite{CelebAMask-HQ}, i.e.
\begin{equation}
		\hat{S}^{19\times 256\times 256} = G_s(\mathbf{F}, \textbf{w}_{s}).
	\label{eq:S}
\end{equation}


\textbf{Semantic-adaptive Domain Translator. }
In addition, we use a semantic-adaptive domain translator, $G_p$, to generate portrait drawings based on features of photographic faces.  
We design $G_p$ following the structure of U-Net \cite{isola2017image}, and use facial semantics $\hat{S}$ to guide synthesis in the manner of SPADE \cite{park2019semantic,Li2021GENRE}. 
Since our model is trained without supervision, the synthesized semantic maps might be incorrect. 
We therefore only use SPADE over small-scale layers in $G_p$ to control the major structure of portrait drawings. 
The process of portrait drawing synthesis is formulated as:
\begin{equation}
	\begin{split}
		\hat{P}^{3\times 256\times 256} = G_p(\hat{I}, \hat{S}).
	\end{split}
	\label{eq:p}
\end{equation}
The semantic guidance will enable our model producing natural and distinct facial structures, as well as semantic-related details (Section \ref{ssec:ablation}). In addition, with the semantic modulation module, our method allows minor editing on portrait drawings (Section \ref{ssec:expapp}). 

\subsection{Two-stage Training}
\label{ssec:2stagetrain}
In preliminary experiments, we train the whole model end-to-end on portrait drawings. The model produces massy ink patterns, and fails to generate acceptable facial structures (Section \ref{ssec:ablation}). 
As previously analyzed, the possible reason is that portrait drawings present sparse appearances with little 3D information.  
To combat this challenge, we propose to train the model in two stages. 
First, we train the feature projector, semantic decoder, and image decoder, to enable them generating high quality facial structures. 
Afterwards, we add the domain translator $G_p$, and refine all the decoding modules for synthesizing multi-view portrait drawings. 
This training strategy conveys well with the architecture of our generator.



\textbf{Training Stage-I.} In the first training stage, we use facial photos and their corresponding semantic maps \cite{CelebAMask-HQ} to train our model without $G_p$. 
We use discriminators similar to pi-GAN \cite{chan2021pi}, and use a semantic discriminator $D_s$ and an image discriminator $D_I$ during training. The loss function of the discriminators are defined as:
\begin{equation}
	\begin{split}
		\mathcal{L}_{D_s} =  &~ \mathbb{E}_{S\sim \mathcal{S}}[f(D_s(S)) + \lambda_1\lvert \nabla D_s(S) \rvert^2] \\ 
		& + \mathbb{E}_{\bm{z}\sim \mathcal{Z},\bm{x}\sim \mathcal{X}}[f(-D_s(\hat{S}))],\\
		\mathcal{L}_{D_I} = &~ \mathbb{E}_{I\sim \mathcal{I}}[f(D_I(I)) + \lambda_1\lvert \nabla D_I(I) \rvert^2] \\ 
		& + \mathbb{E}_{\bm{z}\sim \mathcal{Z},\bm{x}\sim \mathcal{X}}[f(-D_I(\hat{I}))],
\end{split}
\end{equation}
where $f(u) = -\log (1+\exp(-u))$; $I$ and $S$ denote facial photos and semantic maps in the training set $\{ \mathcal{I}, \mathcal{S}\}$. 
The loss function of the generator (without $G_p$) is:
\begin{equation}
	\begin{split}
		\mathcal{L}_{G}^{(1)} = &~ \mathbb{E}_{\bm{z}\sim \mathcal{Z},\bm{x}\sim \mathcal{X}}[f(-D_s(\hat{S})) + f(-D_I(\hat{I}))]+ \lambda_2\mathcal{L}_{rec},\\
		\mathcal{L}_{rec} = &~ \lambda_3\rvert I_{pri}-I_{aux} \lvert+(1-\lambda_3) \mathrm{SSIM}(I_{pri},I_{aux}),
	\end{split}
	\label{eq:Gstage1}
\end{equation}
where $\mathrm{SSIM}$ denotes the structural similarity \cite{SSIM} between $I_1$ and $I_2$; $\lambda_{1,2,3}$ are weighting factors.

\textbf{Training Stage-II.} In the second stage, we load the pre-trained model and add the portrait drawing decoder $G_p$ to it. 
Afterwards, we fix parameters of $G_F$ and refine all the other parts, by using portrait drawings and their semantic maps $\{ P\sim\mathcal{P}, S\sim\mathcal{S} \}$.
Here, we use an portrait drawing discriminator $D_p$ and the previous semantic discriminator $D_s$ during training. 
The loss functions of $D_p$ and $G$ are defined as:
\begin{equation}
\begin{split}
		\mathcal{L}_{D_p} = &~ \mathbb{E}_{P\sim \mathcal{P}}[f(D_p(P)) + \lambda_1\lvert \nabla D_p(P) \rvert^2] \\ 
		& + \mathbb{E}_{\bm{z}\sim \mathcal{Z},\bm{x}\sim \mathcal{X}}[f(-D_p(\hat{P}))], \\
		\mathcal{L}_G^{(2)} = &~ \mathbb{E}_{\bm{z}\sim \mathcal{Z},\bm{x}\sim \mathcal{X}}[f(-D_s(\hat{S})) + f(-D_p(\hat{P}))].
		\end{split}
		\label{eq:loos2}
\end{equation}
We remove the image discriminator $D_I$ and $\mathcal{L}_{rec}$ in this stage, so as to eliminate their interference on synthesizing realistic portrait drawings. 


\begin{figure*}[ht]
	\begin{center}
		\includegraphics[width=1.0\linewidth]{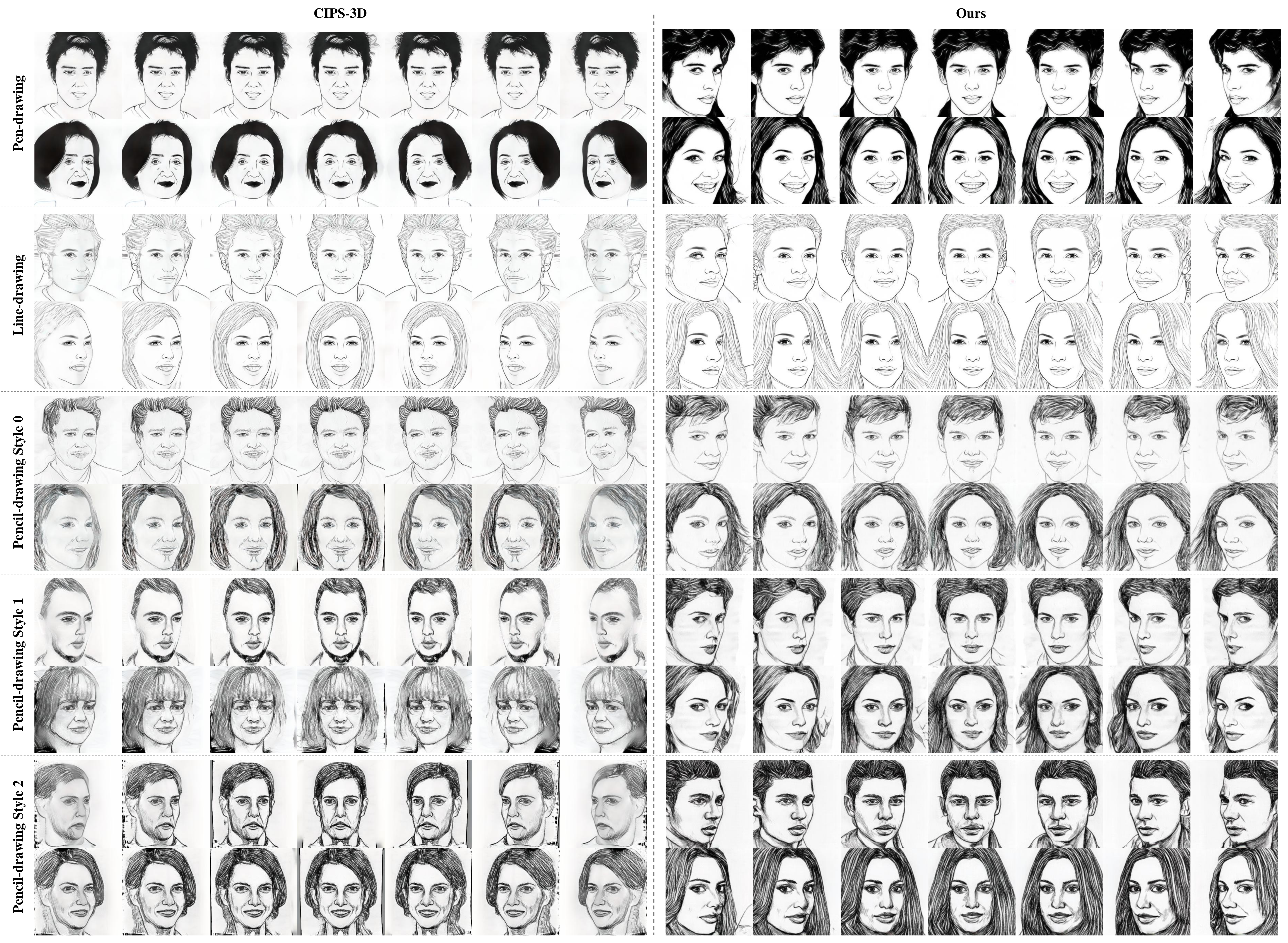}
	\end{center}
	\vspace{-0.3cm}
	\caption{Qualitative comparison on Pen-drawings, Line-drawings, and three styles of Pencil-drawings \protect\cite{Fan2022FS2K}, at $256^2$ resolution. The left part show results of CIPS-3D \protect\cite{zhou2021cips}, the right part show ours.}
	\vspace{-0.2cm}
	\label{fig:compsketch}
\end{figure*}

\subsection{Data Augmentation via Synthesis}
\label{ssec:dataaug}
To effectively learn a 3D-aware image synthesis model, it's critical to collect a huge number of training images of the same style.
This requirement is almost impossible in our task. It's time-consuming and laborious for human artists draw adequate portraits. In existing portrait drawing datasets \cite{Fan2022FS2K,yi2019apdrawinggan}, there are merely hundreds of samples for a given style.
These observations imply the significance of data augmentation for training. In the implementation, we train a modification of GENRE \cite{Li2021GENRE} to synthesize adequate 2D portrait drawings. Afterwards, we use the synthesized data to train our model in Stage-II. 
\begin{figure}[ht]
	\begin{center}
		\includegraphics[width=1.0\linewidth]{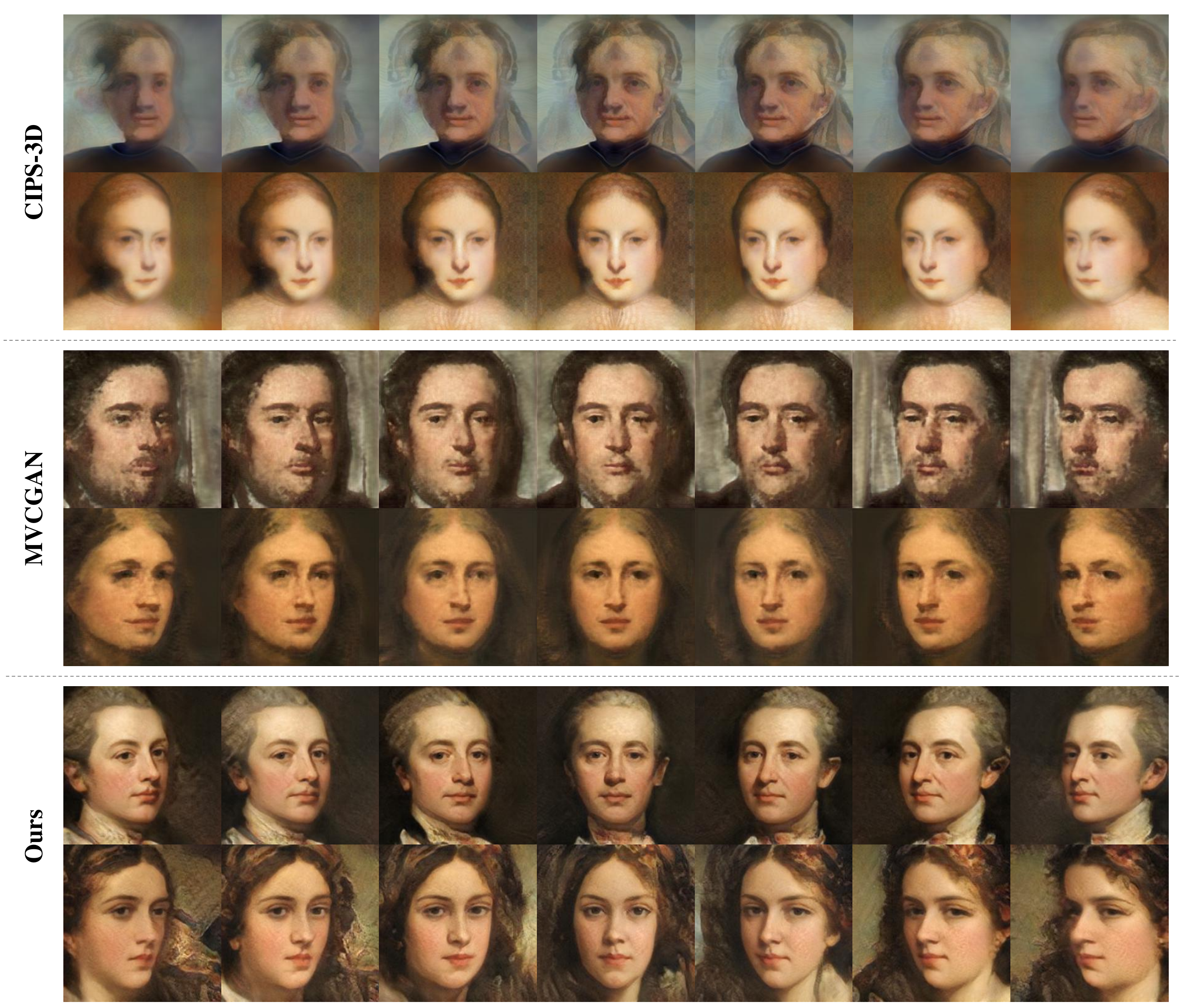}
	\end{center}
	\vspace{-0.3cm}
	\caption{Qualitative comparison in synthesizing multi-view oil-paintings at $256^2$ resolution.}
	\label{fig:compwikiart}
	\vspace{-0.4cm}
\end{figure}

\begin{figure}[h]
	\begin{center}
		\includegraphics[width=1.0\linewidth]{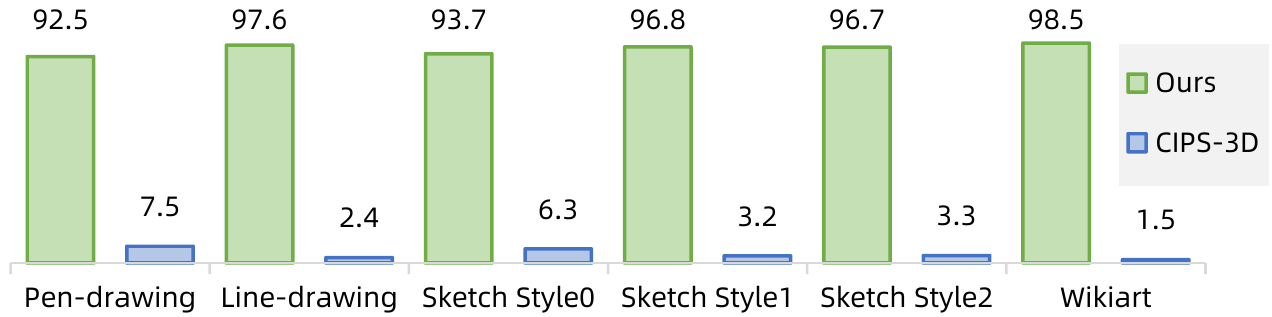}
	\end{center}
	\vspace{-0.3cm}
	\caption{The average preference percent of CIPS-3D and our method w.r.t. each style of artistic portrait drawings.}
	\vspace{-0.4cm}
	\label{fig:subject}
\end{figure}


\section{Experiments and Analysis}
\label{sec:exp}

\subsection{Settings}
\label{ssec:setting}

To verify and analyze our method, we conduct thorough experiments by using the following datasets: \textbf{Pen-drawings.} We conduct experiments on the APDrawing \cite{yi2019apdrawinggan} dataset, which contains 70 pen-drawings of faces. 
\textbf{Line-drawings.} We drew about 800 facial line-drawings ourselves. 
\textbf{Pencil-drawings.} We apply our method to the three styles of pencil-drawings on the FS2K \cite{Fan2022FS2K} dataset. Each style contains about 350 training images. 
\textbf{Oil paintings.} We randomly select 3,133 oil-paintings of humans from WikiArt \cite{nicholpainter} to evaluate the generalization capability of our model.
\textbf{Facial Photos.} We use the CelebAMask-HQ \cite{CelebAMask-HQ} dataset during training stage-I and for data augmentation. It contains 30,000 facial photos and the corresponding semantic maps.
For Pen-drawings, Line-drawings, and three styles of Pencil-drawings, we train a 2D GAN by using the corresponding real portrait drawings. Afterwards, we apply the learned models to the facial photos in CelebAMask-HQ dataset. Finally, the synthesized portrait drawings are used to train 3D-aware image synthesis models.

\subsection{Training Details}
\label{ssec:training}
In the pre-training phase of the model, we set $\lambda_1=0.1$, $\lambda_2=1$, and $\lambda_3=0.25$. In the second phase of training, the parameters remain unchanged. We use the Adam optimizer and set $\beta_1=0$ and $\beta_2=0.9$. At $64^2$ resolution, the learning rate of generator is $6\times 10^{-5}$, the learning rate of discriminators $2\times 10^{-4}$, and batch size 36. At $128^2$ resolution, the learning rate of generator is $5\times 10^{-5}$ and batch size 24. At $256^2$ resolution, the learning rate of generator is $3\times 10^{-5}$, the learning rate of discriminators $1 \times 10^{-4}$, and batch size 24. We use a single GeForce RTX3090 to train our model.

\begin{table*}[ht]
\begin{center}
	\caption{Quantitative comparison between SAGE and advanced methods, including CIPS-3D \protect\cite{zhou2021cips} and MVCGAN \protect\cite{zhang2022mvcgan}. Smaller values of both SIFID and SWD indicate better quality of synthesised images. The best criteria are highlighted in \textbf{bold}.}
	\vspace{-0.2cm}
	\label{tab:comp}
	\scriptsize
	\begin{tabular}{l|cc|cc|cc|cc|cc|cc}
	\toprule	
&\multicolumn{2}{c}{\textit{Pen-drawing}}&\multicolumn{2}{|c}{\textit{Line-drawing}}&\multicolumn{2}{|c}{\textit{Pencil-drawing Style0}}&\multicolumn{2}{|c}{\textit{Pencil-drawing Style1}}&\multicolumn{2}{|c}{\textit{Pencil-drawing Style2}}&\multicolumn{2}{|c}{\textit{Oil-paintings}}\\
	\cmidrule(lr){2-3} \cmidrule(lr){4-5} \cmidrule(lr){6-7} \cmidrule(lr){8-9} \cmidrule(lr){10-11} \cmidrule(lr){12-13}
									&		SIFID		&		SWD		&		SIFID		&		SWD		  &		SIFID		 &		SWD		   &		SIFID		&		SWD		&		SIFID		&		SWD		 &		SIFID		&	SWD		  \\
	\midrule								
	CIPS-3D \cite{zhou2021cips}		&	3.51	&	86.9 	&	3.31	&	91.3 	&	5.49	&	40.7 	&	\textbf{5.12}	&	25.1 	&	5.33	&	24.9 	&	4.86	&	\textbf{14.0} 	\\
	MVCGAN \cite{zhang2022mvcgan}	&	4.19	&	59.8 	&	3.69	&	99.1 	&	5.76	&	48.6 	&	5.15	&	24.7 	&	5.29	&	25.5 	&	4.56	&	36.3 	\\
	SAGE (Ours)	&	\textbf{3.07}	&	\textbf{38.5} 	&	\textbf{3.04}	&	\textbf{89.8} 	&	\textbf{5.16}	&	\textbf{34.0} 	&	5.13	&	\textbf{19.7} 	&	\textbf{5.11}	&	\textbf{19.9} 	&	\textbf{4.71}	&	26.5 	\\
	\bottomrule	
	\end{tabular}
	\vspace{-0.8cm}
\end{center}
\end{table*}


\begin{figure}[t]
	\begin{center}
		\includegraphics[width=1\linewidth]{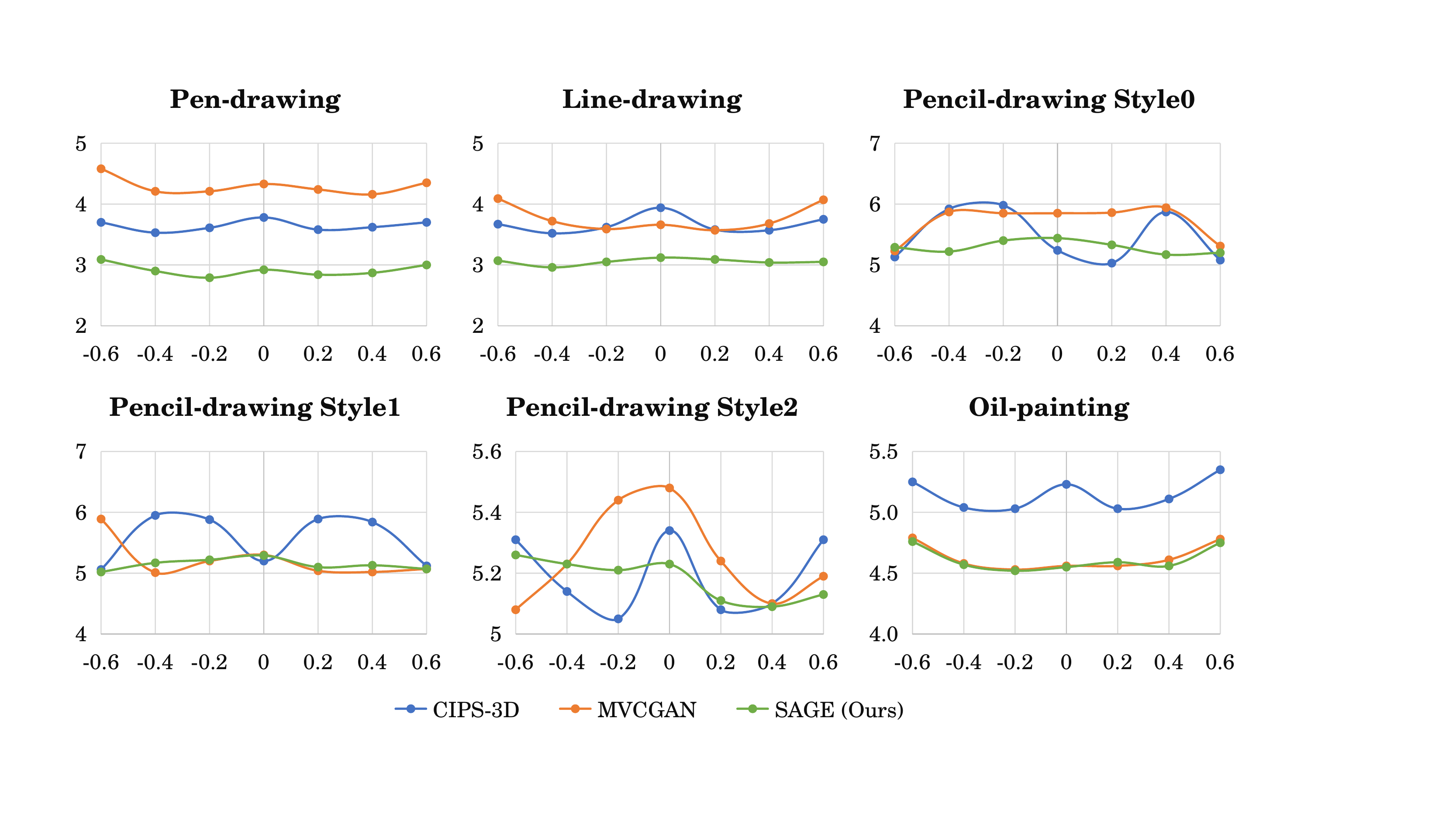}
	\end{center}
	\vspace{-0.3cm}
	\caption{Curves of SIFID values across different views. The $x$-axis denotes the pose variants sequentially corresponding to Fig. \ref{fig:compsketch} and Fig. \ref{fig:compwikiart}. The $y$-axis denotes values of SIFID. } 
	\label{fig:compview}
\end{figure}

\begin{figure}[t]
	\vspace{-0.2cm}
	\begin{center}
		\includegraphics[width=1.0\linewidth]{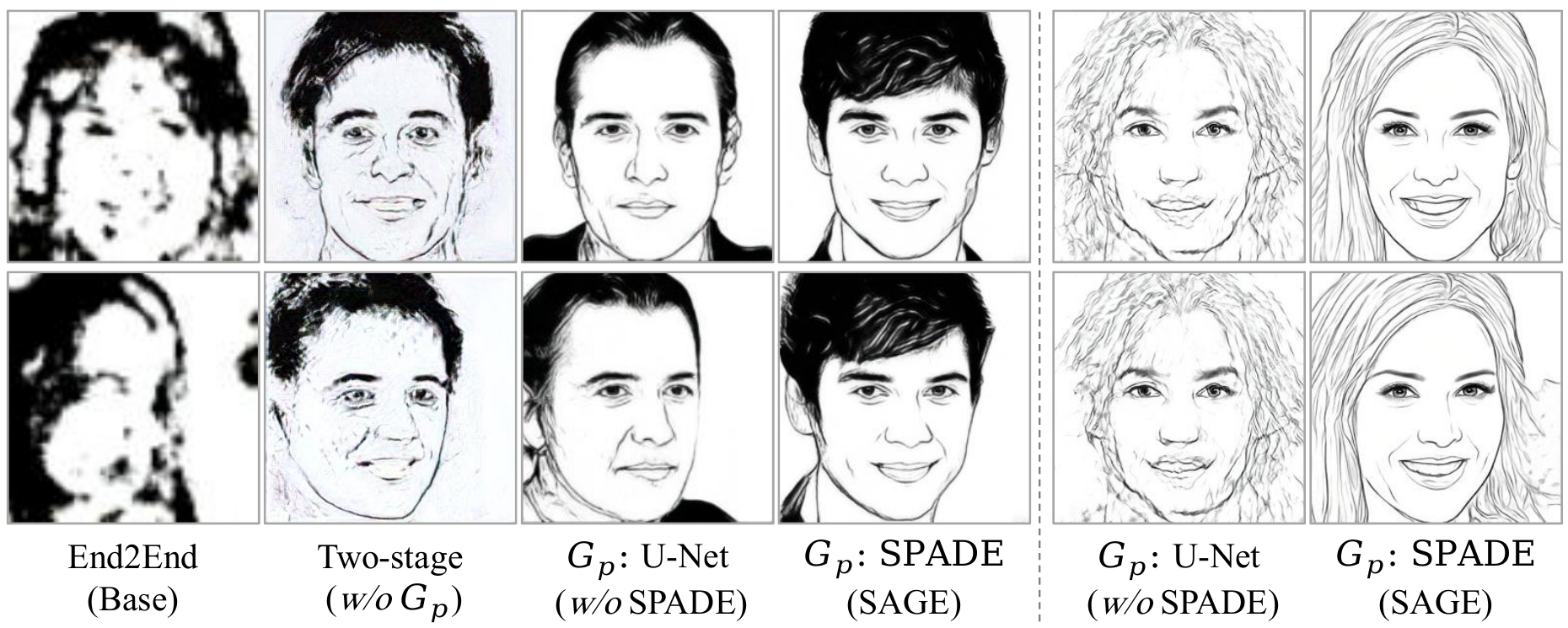}
	\end{center}
	\vspace{-0.3cm}
	\caption{Results of ablation study. 
	}
	\vspace{-0.2cm}
	\label{fig:ablation}
	\vspace{-0.3cm}
\end{figure}

\subsection{Comparison with SOTAs}
\label{ssec:expcomp}
We compare SAGE with two SOTA methods, i.e. CIPS-3D \cite{zhou2021cips} and MVCGAN \cite{zhang2022mvcgan}. We use their official implementations provided by the authors, and conduct experiments under the same settings as ours. 
	
\textbf{Qualitative comparison on portrait-drawings.} As previously presented in Fig. \ref{fig:mtv}, MVCGAN fails to generate acceptable portrait drawings. Fig. \ref{fig:compsketch} shows the portrait drawings generated by CIPS-3D and SAGE. 
Here, we show seven views of each sample. 
Obviously, our results show more natural structures than CIPS-3D. The synthesized drawings of CIPS-3D, especially the pen-drawings and pencil-drawings, present geometric deformations. Besides, CIPS-3D produces unappealing artifacts under large pose variants. In contrast, our method, SAGE, generates high quality portrait drawings across all the poses and all the styles. Specially, our synthesized portrait drawings present distinct facial structures with realistic strokes and textures. 

\textbf{Quantitative comparison on portrait-drawings.} We further evaluate the quality of synthesized images by using the Fr\'{e}chet Inception Distances (FID) \cite{heusel2017gans} and sliced Wasserstein distance (SWD) \cite{karras2017progressive}. For each style of portrait drawing, we use $4K$ generated images and $4K$ real images for computing FID and SWD. As shown in Table\ref{tab:comp}, 
our method achieves significantly lower values of both FID and SWD than CIPS-3D and MVCGAN; except that the SIFID value is slightly higher than CIPS-3D on \textit{Pencil-drawing Style 1}. Such comparison results are consistent with Fig. \ref{fig:compsketch}, and further demonstrate the superiority of our method in generating multi-view portrait drawings.

\textbf{Comparison on oil-paintings.} Since both MVCGAN and CIPS-3D perform well on synthesizing oil-paintings, we additionally compare with them on the WikiArt dataset. 
Fig. \ref{fig:compwikiart} shows the results of multi-view oil-paintings synthesis. Obviously, CIPS-3D produces geometric deformations and unappealing artifacts. MVCGAN produces reasonable paintings but with blurring details. In contrast, our results present the best quality, in terms of either facial structures or painting textures. 
Table \ref{tab:comp} shows the SIFID and SWD values computed from $1500$ generated paintings and $1500$ real ones. Our method achieves the best SIFID value, but shows inferiority to CIPS-3D in terms of SWD. Based on all the comparison results, we can safely conclude that our method is superior and at least highly competitive to SOTAs in synthesizing multi-view oil-paintings. 
	

\textbf{Stability across pose variants.} 
We finally analyze the stability of the three models in generating multi-view images. Specially, we compute the SIFID values corresponding to different viewpoints/poses. Fig. \ref{fig:compview} shows that the SIFID values of either CIPS-3D or MVCGAN present dramatically fluctuate in most cases. Such fluctuations indicate the quality of generated portrait drawings changes dramatically with pose variants. 
In contrast, our method achieves the best and most stable performance in general, across all the six styles. In other words, our method consistently produces high quality multi-view portrait drawings.  
This comparison result is consistent with Fig. \ref{fig:compsketch} and Fig. \ref{fig:compwikiart}. 

\textbf{User study.}
We conducted a series of user study. Specially, we adopt the stimulus-comparison method following ITU-R BT.500-12.
We randomly select 600 generated comparison pairs in total, i.e. 100 pairs for each style.
Each pair is evaluated by 37 participants. 
Finally, we collect 22,200 preference labels. 
Fig.\ref{fig:subject} shows the average preference percent of CIPS-3D and SAGE, w.r.t. each style. 
Obviously, over $92\%$ of participants think the portraits generated by SAGE are better than CIPS-3D, across all the styles. Such results demonstrate that our method is significantly better than CIPS-3D in synthesizing multi-view artistic portrait drawings.

	

\subsection{Ablation Study}
\label{ssec:ablation}


We conduct a series of ablation experiments to validate the significance of our proposed techniques, including two-stage training, the domain translator $G_p$, and the use of semantic masks, i.e. SPADE. We build several model variants by progressively using these techniques on our base model. The experiments are mainly conducted on the APDrawing dataset. The corresponding results are shown in Fig. \ref{fig:ablation} and Table \ref{tab:ablation}.    
	

\textbf{Two-stage training strategy.} We compare end-to-end training (End2End) with two-stage training on the base model. 
As shown in Fig. \ref{fig:ablation}, end-to-end training leads to messy facial structures. In contrast, our two-stage training enables the base model producing natural facial structures. 
Besides, two-stage training dramatically decreases the SIFID and SWD values. 
Such significant performance improvement demonstrate our motivation of using two-stage training. 
	
\textbf{Domain translator $G_p$.} We further evaluate the role of domain translator. We here build a model variant by removing the SPADE modules from $G_p$. the architecture of pristine U-Net. As shown in Fig. \ref{fig:ablation}, using a domain translator, even U-Net, dramatically boosts the quality of generated pen-drawings. 
Correspondingly, Table \ref{tab:ablation} shows that using U-Net alone significantly decreases both SIFID and SWD values. 
These results verify our motivation of using a domain translator to decode portrait drawings from features of facial photos. In this way, our generator can produce distinct facial structures and realism artistic patterns.

\textbf{Semantic guidance.} Finally, we analyze the role of semantics in $G_p$. As shown in Fig.\ref{fig:ablation}, SPADE improves clarity and continuity of facial semantic boundaries. Correspondingly, SPADE further decreases SIFID and SWD; and our full model achieves the best performance among all the model variants. 
To further verify the significance of facial semantics, we conduct experiments on line-drawing synthesis. The last two columns of Fig.\ref{fig:ablation} show the corresponding results. Obviously, without the guidance of facial semantics, the model generate chaotic lines. In contrast, our full model generate distinctly better line-drawings. 

These results demonstrate that semantic guidance are crucial for generating multi-style portrait drawings.  From one hand, semantic masks enable the domain translator producing distinct facial structures. For the other hand, facial semantics are highly correlated with the drawing techniques, human artists used during the creation process.

\begin{table}[t]
	\centering
	\caption{Quantitative results of ablation experiments about the training strategy and the domain translator $G_p$, on pen-drawing synthesis.}
	\label{tab:ablation}
	\vspace{-0.2cm}
	\scriptsize
	\begin{tabular}{cc|cc|cc}
		\toprule
		\multicolumn{2}{c|}{\textit{Training}}&\multicolumn{2}{c|}{$G_p$}	&			&\\
		\cmidrule(lr){1-2} \cmidrule(lr){3-4}
		End2End		&	Two-Stage	 &		U-Net		&		SPADE		&	SIFID	&	SWD	\\
		\midrule
		\ding{52}	&				 &					&					&	7.36	&	92.78	\\
		&	\ding{52}	 &					&					&	4.29	&	65.83	\\
		&	\ding{52}	 &	\ding{52}		&					&	3.18	&	43.27	\\
		&	\ding{52}	 &	\ding{52}		&	\ding{52}		&\textbf{3.07}&	\textbf{38.45}	\\
		\bottomrule		
	\end{tabular}
	\vspace{-0.3cm}
\end{table}

\begin{figure}[t]
	\begin{center}
		\includegraphics[width=1\linewidth]{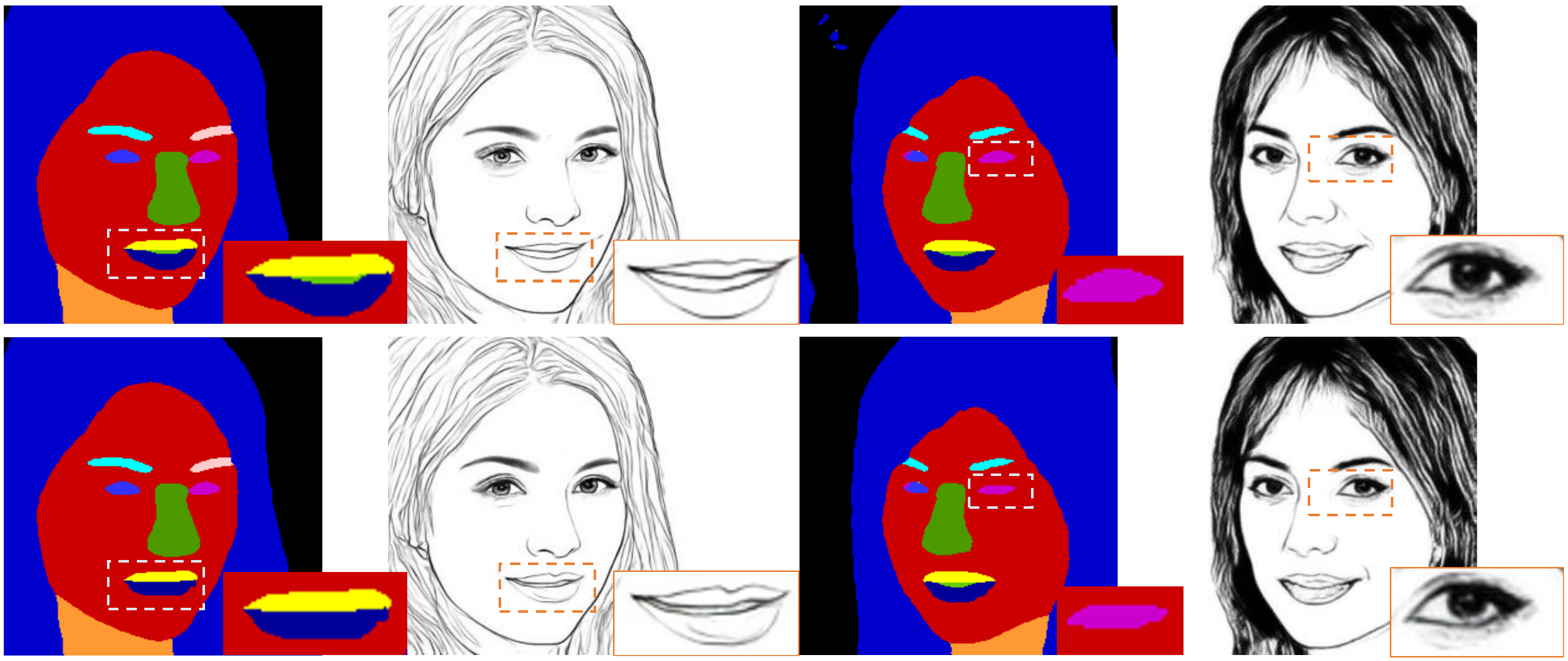}
	\end{center}
	\vspace{-0.3cm}
	\caption{Semantic editing on synthesized portrait drawings.}
	\vspace{-0.4cm}
	\label{fig:edit}
\end{figure}
%


\begin{figure}[t]
	\begin{center}
		\includegraphics[width=1.0\linewidth]{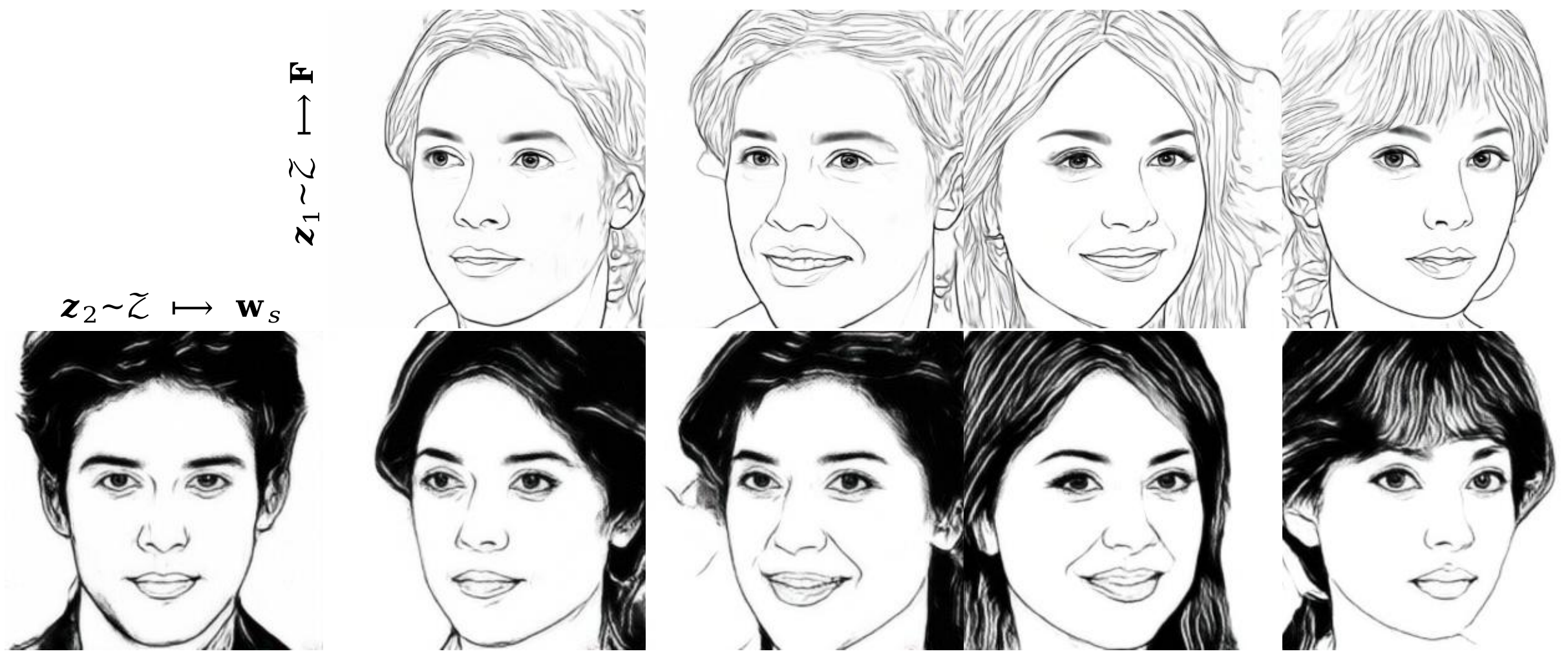}
	\end{center}
	\vspace{-0.3cm}
	\caption{Style transfer from line-drawings to pen-drawings. 
	}
	\vspace{-0.4cm}
	\label{fig:stylemix}
\end{figure}
\begin{figure}[h]
	\begin{center}
		\includegraphics[width=1.0\linewidth]{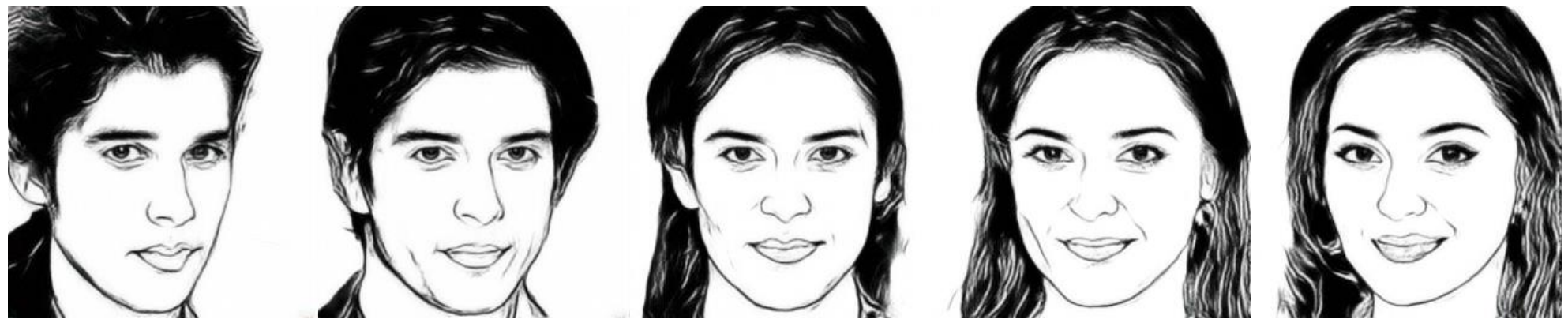}
	\end{center}
	\vspace{-0.3cm}
	\caption{Result of identity interpolation. 
	}
	\vspace{-0.2cm}
	\label{fig:idinter}
\end{figure}

\subsection{Applications}
\label{ssec:expapp}
\textbf{Semantic editing.} Recall that we use semantic maps in the domain translator $G_p$ to control facial structures of portrait drawings. As a result, our model allows for editing of portrait drawings, to a certain extent. In other words, if we modify the input semantics of $G_p$ slightly, the synthesized portrait drawing will change accordingly. 
As shown in Fig. \ref{fig:edit}, when we change the semantic masks of teeth or eyes, the corresponding areas in synthesized images change accordingly. 

\textbf{Style transfer.} In our generator, the content and style information are disentangled and stored in $\mathbf{F}$ and $\mathbf{w}_s$, respectively. It is thus possible for us to change a portrait drawing to another style without changing the facial content. To this end, we put a latent code ${\mathbf{z}_1}$ into the learned \textit{line-drawing} model to get the content feature $\mathbf{F}$; and put another latent code ${\mathbf{z}_2}$ into the learned \textit{pen-drawing} model to get the style vector $\mathbf{w}_s$. Afterwards, we use $\mathbf{F}$ and $\mathbf{w}_s$ for decoding a \textit{pen-drawing}. As shown in Fig. \ref{fig:stylemix}, the line-drawings are transferred to pen-drawings, while preserving facial semantic content. 	

\textbf{Identity interpolation.} We also perform identity interpolation experiments on SAGE. Given two synthesized images, we perform linear interpolation in the content space $\textbf{w}_c \sim \mathcal{W}_c$ and viewpoint space $\mathbf{x} \sim \mathcal{X}$. Fig. \ref{fig:idinter} shows interpolation results on Pen-drawings. The smooth transition in pose and appearance implies that SAGE allows precise control on both facial identity and pose. 


	
%

	
\section{Conclusion}
\label{ssec:conclusion}
We propose a novel method, SAGE, for generating multi-view portrait drawings. 
SAGE is designed to collaboratively synthesize facial photos, semantic masks, and portrait drawings. 
Extensive experiments and a series of ablation study are conducted on six styles of artistic portraits. SAGE stably shows impressive performance in generating high quality portrait drawings, and outperforms previous 3D-aware image synthesis methods. In the future, we are interested in synthesizing multi-view portrait drawings conditioned on a single photo. One possible way is to incorporate GAN-inversion and few-shot learning techniques. 

\bibliographystyle{named}
\bibliography{ref}

\begin{thebibliography}{}

\bibitem[\protect\citeauthoryear{Chan \bgroup \em et al.\egroup
  }{2021}]{chan2021pi}
Eric~R Chan, Marco Monteiro, Petr Kellnhofer, Jiajun Wu, and Gordon Wetzstein.
\newblock pi-gan: Periodic implicit generative adversarial networks for
  3d-aware image synthesis.
\newblock In {\em Proceedings of the IEEE/CVF conference on computer vision and
  pattern recognition}, pages 5799--5809, 2021.

\bibitem[\protect\citeauthoryear{Chan \bgroup \em et al.\egroup
  }{2022a}]{chan2022informativedraw}
Caroline Chan, Fr{\'e}do Durand, and Phillip Isola.
\newblock Learning to generate line drawings that convey geometry and
  semantics.
\newblock In {\em Proceedings of the IEEE/CVF Conference on Computer Vision and
  Pattern Recognition}, pages 7915--7925, 2022.

\bibitem[\protect\citeauthoryear{Chan \bgroup \em et al.\egroup
  }{2022b}]{chan2022efficient}
Eric~R Chan, Connor~Z Lin, Matthew~A Chan, Koki Nagano, Boxiao Pan, Shalini
  De~Mello, Orazio Gallo, Leonidas~J Guibas, Jonathan Tremblay, Sameh Khamis,
  et~al.
\newblock Efficient geometry-aware 3d generative adversarial networks.
\newblock In {\em Proceedings of the IEEE/CVF Conference on Computer Vision and
  Pattern Recognition}, pages 16123--16133, 2022.

\bibitem[\protect\citeauthoryear{Deng \bgroup \em et al.\egroup
  }{2022a}]{deng2022depth}
Kangle Deng, Andrew Liu, Jun-Yan Zhu, and Deva Ramanan.
\newblock Depth-supervised nerf: Fewer views and faster training for free.
\newblock In {\em Proceedings of the IEEE/CVF Conference on Computer Vision and
  Pattern Recognition}, pages 12882--12891, 2022.

\bibitem[\protect\citeauthoryear{Deng \bgroup \em et al.\egroup
  }{2022b}]{deng2022gram}
Yu~Deng, Jiaolong Yang, Jianfeng Xiang, and Xin Tong.
\newblock Gram: Generative radiance manifolds for 3d-aware image generation.
\newblock In {\em Proceedings of the IEEE/CVF Conference on Computer Vision and
  Pattern Recognition}, pages 10673--10683, 2022.

\bibitem[\protect\citeauthoryear{Fan \bgroup \em et al.\egroup
  }{2022}]{Fan2022FS2K}
Deng-Ping Fan, Ziling Huang, Peng Zheng, Hong Liu, Xuebin Qin, and Luc
  Van~Gool.
\newblock Facial-sketch synthesis: A new challenge.
\newblock {\em Machine Intelligence Research}, 19(4):257--287, 2022.

\bibitem[\protect\citeauthoryear{Gao \bgroup \em et al.\egroup
  }{2020}]{gao2020making}
Fei Gao, Jingjie Zhu, Zeyuan Yu, Peng Li, and Tao Wang.
\newblock Making robots draw a vivid portrait in two minutes.
\newblock In {\em 2020 IEEE/RSJ International Conference on Intelligent Robots
  and Systems (IROS)}, pages 9585--9591. IEEE, 2020.

\bibitem[\protect\citeauthoryear{Gu \bgroup \em et al.\egroup
  }{2021}]{gu2021stylenerf}
Jiatao Gu, Lingjie Liu, Peng Wang, and Christian Theobalt.
\newblock Stylenerf: A style-based 3d-aware generator for high-resolution image
  synthesis.
\newblock {\em arXiv preprint arXiv:2110.08985}, 2021.

\bibitem[\protect\citeauthoryear{Gui \bgroup \em et al.\egroup
  }{2021}]{Gui2020ReviewGAN}
J~Gui, Z~Sun, Y.~Wen, D.~Tao, and J.~Ye.
\newblock A review on generative adversarial networks: Algorithms, theory, and
  applications.
\newblock {\em IEEE Transactions on Knowledge and Data Engineering}, pages
  1--28, 2021.

\bibitem[\protect\citeauthoryear{Heusel \bgroup \em et al.\egroup
  }{2017}]{heusel2017gans}
Martin Heusel, Hubert Ramsauer, Thomas Unterthiner, Bernhard Nessler, and Sepp
  Hochreiter.
\newblock Gans trained by a two time-scale update rule converge to a local nash
  equilibrium.
\newblock {\em Advances in neural information processing systems}, 30, 2017.

\bibitem[\protect\citeauthoryear{Huang and Belongie}{2017}]{huang2017arbitrary}
Xun Huang and Serge Belongie.
\newblock Arbitrary style transfer in real-time with adaptive instance
  normalization.
\newblock In {\em Proceedings of the IEEE international conference on computer
  vision}, pages 1501--1510, 2017.

\bibitem[\protect\citeauthoryear{Isola \bgroup \em et al.\egroup
  }{2017}]{isola2017image}
Phillip Isola, Jun-Yan Zhu, Tinghui Zhou, and Alexei~A Efros.
\newblock Image-to-image translation with conditional adversarial networks.
\newblock In {\em Proceedings of the IEEE conference on computer vision and
  pattern recognition}, pages 1125--1134, 2017.

\bibitem[\protect\citeauthoryear{Karras \bgroup \em et al.\egroup
  }{2017}]{karras2017progressive}
Tero Karras, Timo Aila, Samuli Laine, and Jaakko Lehtinen.
\newblock Progressive growing of gans for improved quality, stability, and
  variation.
\newblock {\em arXiv preprint arXiv:1710.10196}, 2017.

\bibitem[\protect\citeauthoryear{Karras \bgroup \em et al.\egroup
  }{2019}]{karras2019style}
Tero Karras, Samuli Laine, and Timo Aila.
\newblock A style-based generator architecture for generative adversarial
  networks.
\newblock In {\em 2019 IEEE/CVF Conference on Computer Vision and Pattern
  Recognition (CVPR)}, pages 4396--4405. IEEE, 2019.

\bibitem[\protect\citeauthoryear{Kong \bgroup \em et al.\egroup
  }{2021}]{kong2021smoothing}
Chaerin Kong, Jeesoo Kim, Donghoon Han, and Nojun Kwak.
\newblock Smoothing the generative latent space with mixup-based distance
  learning.
\newblock {\em arXiv preprint arXiv:2111.11672}, 2021.

\bibitem[\protect\citeauthoryear{Lee \bgroup \em et al.\egroup
  }{2020}]{CelebAMask-HQ}
Cheng-Han Lee, Ziwei Liu, Lingyun Wu, and Ping Luo.
\newblock Maskgan: Towards diverse and interactive facial image manipulation.
\newblock In {\em IEEE Conference on Computer Vision and Pattern Recognition
  (CVPR)}, 2020.

\bibitem[\protect\citeauthoryear{Li \bgroup \em et al.\egroup
  }{2020}]{li2020few}
Yijun Li, Richard Zhang, Jingwan Lu, and Eli Shechtman.
\newblock Few-shot image generation with elastic weight consolidation.
\newblock {\em arXiv preprint arXiv:2012.02780}, 2020.

\bibitem[\protect\citeauthoryear{Li \bgroup \em et al.\egroup
  }{2021}]{Li2021GENRE}
Xiang Li, Fei Gao, and Fei Huang.
\newblock High-quality face sketch synthesis via geometric normalization and
  regularization.
\newblock In {\em IEEE International Conference on Multimedia and Expo (ICME)
  2021}, July 5-9 2021.

\bibitem[\protect\citeauthoryear{Mildenhall \bgroup \em et al.\egroup
  }{2021}]{mildenhall2021nerf}
Ben Mildenhall, Pratul~P Srinivasan, Matthew Tancik, Jonathan~T Barron, Ravi
  Ramamoorthi, and Ren Ng.
\newblock Nerf: Representing scenes as neural radiance fields for view
  synthesis.
\newblock {\em Communications of the ACM}, 65(1):99--106, 2021.

\bibitem[\protect\citeauthoryear{Nichol}{2016}]{nicholpainter}
K~Nichol.
\newblock Painter by numbers, wikiart (2016).
\newblock {\em URL https://www. kaggle. com/c/painter-by-numbers}, 2016.

\bibitem[\protect\citeauthoryear{Niemeyer and
  Geiger}{2021}]{niemeyer2021giraffe}
Michael Niemeyer and Andreas Geiger.
\newblock Giraffe: Representing scenes as compositional generative neural
  feature fields.
\newblock In {\em Proceedings of the IEEE/CVF Conference on Computer Vision and
  Pattern Recognition}, pages 11453--11464, 2021.

\bibitem[\protect\citeauthoryear{Noguchi and Harada}{2019}]{noguchi2019image}
Atsuhiro Noguchi and Tatsuya Harada.
\newblock Image generation from small datasets via batch statistics adaptation.
\newblock In {\em Proceedings of the IEEE/CVF International Conference on
  Computer Vision}, pages 2750--2758, 2019.

\bibitem[\protect\citeauthoryear{Ojha \bgroup \em et al.\egroup
  }{2021}]{ojha2021few}
Utkarsh Ojha, Yijun Li, Jingwan Lu, Alexei~A Efros, Yong~Jae Lee, Eli
  Shechtman, and Richard Zhang.
\newblock Few-shot image generation via cross-domain correspondence.
\newblock In {\em Proceedings of the IEEE/CVF Conference on Computer Vision and
  Pattern Recognition}, pages 10743--10752, 2021.

\bibitem[\protect\citeauthoryear{Or-El \bgroup \em et al.\egroup
  }{2022}]{or2022stylesdf}
Roy Or-El, Xuan Luo, Mengyi Shan, Eli Shechtman, Jeong~Joon Park, and Ira
  Kemelmacher-Shlizerman.
\newblock Stylesdf: High-resolution 3d-consistent image and geometry
  generation.
\newblock In {\em Proceedings of the IEEE/CVF Conference on Computer Vision and
  Pattern Recognition}, pages 13503--13513, 2022.

\bibitem[\protect\citeauthoryear{Park \bgroup \em et al.\egroup
  }{2019}]{park2019semantic}
Taesung Park, Ming-Yu Liu, Ting-Chun Wang, and Jun-Yan Zhu.
\newblock Semantic image synthesis with spatially-adaptive normalization.
\newblock In {\em Proceedings of the IEEE/CVF conference on computer vision and
  pattern recognition}, pages 2337--2346, 2019.

\bibitem[\protect\citeauthoryear{Perez \bgroup \em et al.\egroup
  }{2018}]{perez2018film}
Ethan Perez, Florian Strub, Harm De~Vries, Vincent Dumoulin, and Aaron
  Courville.
\newblock Film: Visual reasoning with a general conditioning layer.
\newblock In {\em Proceedings of the AAAI Conference on Artificial
  Intelligence}, volume~32, 2018.

\bibitem[\protect\citeauthoryear{Schwarz \bgroup \em et al.\egroup
  }{2020}]{schwarz2020graf}
Katja Schwarz, Yiyi Liao, Michael Niemeyer, and Andreas Geiger.
\newblock Graf: Generative radiance fields for 3d-aware image synthesis.
\newblock {\em Advances in Neural Information Processing Systems},
  33:20154--20166, 2020.

\bibitem[\protect\citeauthoryear{Sousa and Buchanan}{1999}]{sousa1999pencil}
Mario~Costa Sousa and John~W Buchanan.
\newblock Computer-generated graphite pencil rendering of 3d polygonal models.
\newblock In {\em Computer Graphics Forum}, volume~18, pages 195--208. Wiley
  Online Library, 1999.

\bibitem[\protect\citeauthoryear{Sun \bgroup \em et al.\egroup
  }{2022a}]{sun2022ide}
Jingxiang Sun, Xuan Wang, Yichun Shi, Lizhen Wang, Jue Wang, and Yebin Liu.
\newblock Ide-3d: Interactive disentangled editing for high-resolution 3d-aware
  portrait synthesis.
\newblock {\em arXiv e-prints}, pages arXiv--2205, 2022.

\bibitem[\protect\citeauthoryear{Sun \bgroup \em et al.\egroup
  }{2022b}]{sun2022fenerf}
Jingxiang Sun, Xuan Wang, Yong Zhang, Xiaoyu Li, Qi~Zhang, Yebin Liu, and Jue
  Wang.
\newblock Fenerf: Face editing in neural radiance fields.
\newblock In {\em Proceedings of the IEEE/CVF Conference on Computer Vision and
  Pattern Recognition}, pages 7672--7682, 2022.

\bibitem[\protect\citeauthoryear{Wang and Tang}{2009}]{Wang2009Face}
X.~Wang and X.~Tang.
\newblock Face photo-sketch synthesis and recognition.
\newblock {\em IEEE TPAMI}, 31(11):1955--1967, 2009.

\bibitem[\protect\citeauthoryear{Wang \bgroup \em et al.\egroup }{2004}]{SSIM}
Zhou Wang, A.C. Bovik, H.R. Sheikh, and E.P. Simoncelli.
\newblock Image quality assessment: from error visibility to structural
  similarity.
\newblock {\em IEEE Transactions on Image Processing}, 13(4):600--612, 2004.

\bibitem[\protect\citeauthoryear{Wang \bgroup \em et al.\egroup
  }{2017}]{Wang2017Face}
N.~Wang, S.~Zhang, C.~Peng, J.~Li, and X.~Gao.
\newblock {\em Face Sketch Recognition via Data-Driven Synthesis}.
\newblock 2017.

\bibitem[\protect\citeauthoryear{Wang \bgroup \em et al.\egroup
  }{2018}]{wang2018high}
Ting-Chun Wang, Ming-Yu Liu, Jun-Yan Zhu, Andrew Tao, Jan Kautz, and Bryan
  Catanzaro.
\newblock High-resolution image synthesis and semantic manipulation with
  conditional gans.
\newblock In {\em Proceedings of the IEEE conference on computer vision and
  pattern recognition}, pages 8798--8807, 2018.

\bibitem[\protect\citeauthoryear{Xiang \bgroup \em et al.\egroup
  }{2022}]{xiang2022gram}
Jianfeng Xiang, Jiaolong Yang, Yu~Deng, and Xin Tong.
\newblock Gram-hd: 3d-consistent image generation at high resolution with
  generative radiance manifolds.
\newblock {\em arXiv preprint arXiv:2206.07255}, 2022.

\bibitem[\protect\citeauthoryear{Yi \bgroup \em et al.\egroup
  }{2019}]{yi2019apdrawinggan}
Ran Yi, Yong-Jin Liu, Yu-Kun Lai, and Paul~L Rosin.
\newblock Apdrawinggan: Generating artistic portrait drawings from face photos
  with hierarchical gans.
\newblock In {\em Proceedings of the IEEE/CVF Conference on Computer Vision and
  Pattern Recognition}, pages 10743--10752, 2019.

\bibitem[\protect\citeauthoryear{Yi \bgroup \em et al.\egroup }{2022}]{YiLLR22}
Ran Yi, Yong-Jin Liu, Yu-Kun Lai, and Paul~L Rosin.
\newblock Quality metric guided portrait line drawing generation from unpaired
  training data.
\newblock {\em IEEE Transactions on Pattern Analysis and Machine Intelligence},
  45:905--918, 2022.

\bibitem[\protect\citeauthoryear{Yu \bgroup \em et al.\egroup
  }{2021}]{gao2021cagan}
J.~Yu, S.~Shi, F.~Gao, D.~Tao, and Q.~Huang.
\newblock Towards realistic face photo-sketch synthesis via composition-aided
  {GANs}.
\newblock {\em IEEE TCYB}, 51(9):4350--4362, 2021.

\bibitem[\protect\citeauthoryear{Zhang \bgroup \em et al.\egroup
  }{2018}]{Zhang2018IJCAI}
S.~Zhang, R.~Ji, J.~Hu, Y.~Gao, and Lin C.-W.
\newblock Robust face sketch synthesis via generative adversarial fusion of
  priors and parametric sigmoid.
\newblock In {\em IJCAI}, pages 1163--1169, 2018.

\bibitem[\protect\citeauthoryear{{Zhang} \bgroup \em et al.\egroup
  }{2019}]{Zhang2019TIP}
M.~{Zhang}, R.~{Wang}, X.~{Gao}, J.~{Li}, and D.~{Tao}.
\newblock Dual-transfer face sketch–photo synthesis.
\newblock {\em TIP}, 28(2):642--657, Feb 2019.

\bibitem[\protect\citeauthoryear{Zhang \bgroup \em et al.\egroup
  }{2022}]{zhang2022mvcgan}
Xuanmeng Zhang, Zhedong Zheng, Daiheng Gao, Bang Zhang, Pan Pan, and Yi~Yang.
\newblock Multi-view consistent generative adversarial networks for 3d-aware
  image synthesis.
\newblock In {\em Proceedings of the IEEE/CVF Conference on Computer Vision and
  Pattern Recognition}, pages 18450--18459, 2022.

\bibitem[\protect\citeauthoryear{Zhou \bgroup \em et al.\egroup
  }{2021}]{zhou2021cips}
Peng Zhou, Lingxi Xie, Bingbing Ni, and Qi~Tian.
\newblock Cips-3d: A 3d-aware generator of gans based on
  conditionally-independent pixel synthesis.
\newblock {\em arXiv e-prints}, pages arXiv--2110, 2021.

\bibitem[\protect\citeauthoryear{Zhu \bgroup \em et al.\egroup
  }{2021}]{zhu2021sketch}
Mingrui Zhu, Changcheng Liang, Nannan Wang, Xiaoyu Wang, Zhifeng Li, and Xinbo
  Gao.
\newblock A sketch-transformer network for face photo-sketch synthesis.
\newblock In {\em International Joint Conference on Artificial Intelligence},
  2021.

\end{thebibliography}

\end{document}